\documentclass[10pt,twocolumn,letterpaper]{article}
\usepackage{amsmath}
\usepackage{amsfonts}
\usepackage[dvipsnames]{xcolor}
\usepackage{tabularx, booktabs}
\usepackage{gensymb}
\usepackage{multirow}
\usepackage{tabularx, booktabs}
\usepackage[pagebackref,breaklinks,colorlinks,citecolor=cvprblue]{hyperref}
\usepackage{subfigure}
\usepackage{graphicx}

\definecolor{cvprblue}{rgb}{0.21,0.49,0.74}

\input{tcilatex}
\begin{document}

\title{Estimating Extreme 3D Image Rotations using Cascaded Attention }
\author{Shay Dekel \\
Bar Ilan University\\
Ramat-Gan, Israel\\
\texttt{\small shaydekel@gmail.com} \and Yosi Keller \\
Bar Ilan University\\
Ramat-Gan, Israel\\
\texttt{\small yosi.keller@gmail.com}  \and Martin \v{C}ad\'{i}k \\
FIT, Brno University of Technology\\
Brno, Czech Republic\\
\texttt{\small cadik@fit.vut.cz}
}
\maketitle

\begin{abstract}
Estimating large, extreme inter-image rotations is critical for numerous
computer vision domains involving images related by limited or
non-overlapping fields of view. In this work, we propose an attention-based
approach with a pipeline of novel algorithmic components. First, as rotation
estimation pertains to image pairs, we introduce an inter-image distillation
scheme using Decoders to improve embeddings. Second, whereas contemporary
methods compute a 4D correlation volume (4DCV) encoding inter-image
relationships, we propose an Encoder-based cross-attention approach between
activation maps to compute an enhanced equivalent of the 4DCV. Finally, we
present a cascaded Decoder-based technique for alternately refining the
cross-attention and the rotation query. Our approach outperforms current
state-of-the-art methods on extreme rotation estimation. We make our code
publicly available\QQfnmark{{\scriptsize %
\url{https:/anonymous.4open.science/r/AttExtremeRotation-A467/}}}.
\end{abstract}

\section{Introduction}

\label{intro}

\begin{figure}[tbh]
\centering
\subfigure[]{{\includegraphics[width=3.5cm]{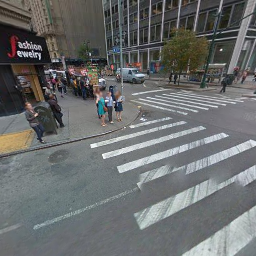} }} %
\subfigure[]{{\includegraphics[width=3.5cm]{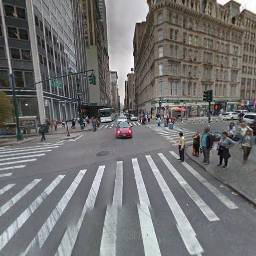} }} \\
\subfigure[]{{\includegraphics[width=3.5cm]{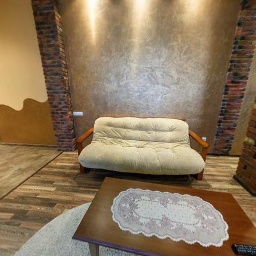} }} %
\subfigure[]{{\includegraphics[width=3.5cm]{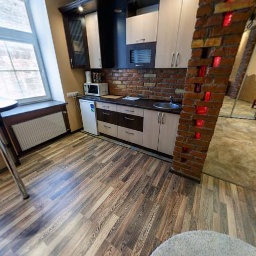} }}
\caption{The estimation of extreme 3D image rotations. First row: Images pair with a small overlap. Second row: non-overlapping image pairs. The proposed scheme estimates the relative rotation between image pairs. }
\label{fig:example}
\end{figure}

Estimating the relative pose between a pair of images is a crucial task in
computer vision, which is used in various applications such as indoor
navigation, augmented reality, autonomous driving, 3D reconstruction \cite%
{schonberger2016structure,ozyesil2017survey}, camera localization \cite%
{brachmann2017dsac,schonberger2018semantic,10.1007/978-3-031-20080-9_9}, simultaneous
localization and mapping \cite{davison2007monoslam,mur2015orb}, and novel
view synthesis \cite{mildenhall2020nerf, riegler2020fvs}. The current
approach to image registration involves extracting features, matching them,
and establishing correspondence between them. However, this approach is
ineffective for input pairs with little or no overlap, making it difficult
to establish sufficient feature correspondences for matching, such as in the
images shown in Fig. \ref{fig:example}.

Numerous applications \cite{PhotoSequencing, ApartmentsCVPR15,
elor2015ringit} necessitate precise estimation of inter-image rotations. The
prevalent approach for extreme 3D rotation estimation between images with
limited or no overlap, as in Fig. \ref{fig:example}, relates to the seminal
work of Coughlan and Yuille \cite{790349}. They introduced a technique
premised on linear structures within an image, primarily arising from three
mutually orthogonal directions - one vertical (building walls) and two
horizontal (ground pavements, roads, etc.). Similarly, "Single View
Metrology" by Criminisi et al. \cite{DBLP:conf/iccv/CriminisiRZ99} and
extensions \cite%
{10.1007/978-3-030-58621-8_19,10.1007/978-3-030-58610-2_32,10.1007/978-3-031-19769-7_15}
utilize parallel image lines and corresponding vanishing points \cite%
{Hartley2004} for camera calibration. Furthermore, relative camera rotation
can be estimated via illumination cues \cite{4409005}, by analyzing lighting
and shadow directions.

\begin{figure*}[tbh]
\centering\includegraphics[width=\textwidth]{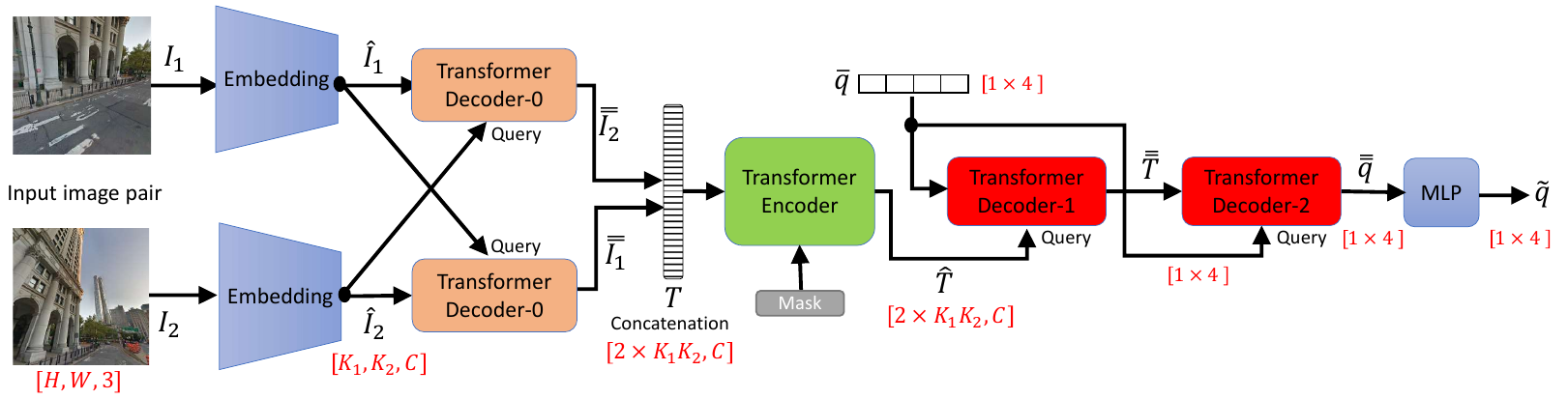}
\caption{\textbf{The proposed architecture} utilizes weight-sharing
Siamese CNNs to encode the input image pair $%
(I_{1},I_{2})\in \mathbb{R}^{H\times W}$ into feature maps $(\hat{I}_{1},%
\hat{I}_{2})$. These feature maps are then cross-decoded by the weight
sharing Transformer Decoder-0 layers, cross-distilling $(\hat{I}_{1},\hat{I}%
_{2})$ into the representations $\Bar{\Bar{I}}_{1}$ and $\Bar{\Bar{I}}_{2}$.
The concatenated refined embeddings $T$ are input to the Transformer-Encoder
alongside an attention mask $M$ to derive the cross-attention encoding $\hat{%
T}$. $\hat{T}$ enters a cascade of two Transformer Decoders, where the
first, Transformer Decoder-1, enhances the cross-attention as $\Bar{\Bar{T}}$%
, guided by the learned quaternion rotation query $\bar{q}$. The second,
Transformer Decoder-2, encodes the rotation as $\Bar{\Bar{q}}$, transformed
via a multilayer perceptron (MLP) to predict the relative quaternion
rotation $\Tilde{q}$.}
\label{fig:overview}
\vspace{-8pt}
\end{figure*}

In this work, we propose a deep-learning approach for estimating
significant, extreme inter-image rotations. Unlike classical formulations
\cite{790349,DBLP:conf/iccv/CriminisiRZ99} that explicitly detect
hand-crafted cues such as lines, shadows, and vanishing points, our method
directly regresses the relative rotation from input images through a deep neural
network. Inspired by recent successful applications of Transformers \cite%
{NIPS2017_3f5ee243} in computer vision tasks including object detection \cite%
{ObjectDetectionTransformers} and image recognition \cite%
{TransformersImageRecognition}, we adapt Transformers for \textit{multiple}
tasks within the proposed pipeline shown in Fig.~\ref{fig:overview},
expanding beyond previous applications of Transformers.

First, we apply Transformers-Decoders to improve the input image embeddings
by distilling inter-image information between the images by cross-decoding,
where each embedding uses the other's embedding as a query. This better
encodes images with respect to each other. Second, a Transformer-Encoder
computes a stacked multihead attention to encode \textit{cross-attention}
between the latent representations of image pairs. Thus, it improves on the
4D correlation volume (4DCV) used in prior works \cite%
{RAFT,nie2019multi,liang2019stereo,9495134,dosovitskiy2015flownet}, where
4DCVs were calculated by inner products. Instead of a single layer of $N^{2}$
inner products as in 4DCV, the proposed Transformer-Encoder-based approach
leverages multi-head attention's advanced architecture to better encode
interactions between activation map entries. Third, we further improve the
cross-attention encoding using a cascade of two decoders and a learnt
rotation query, to jointly refine the cross-attention encoding and the
rotation query. The proposed scheme is a \textit{ general-purpose}
attention-based architecture for estimating attributes related to two input
images such as optical flow, registration, relative pose regression, etc.
This work was motivated by extreme rotation estimation, and we reserve other
applications for future work, as those will require additional task-specific
modifications.

Interestingly, the attention maps computed by our scheme, shown in Section %
\ref{subsec:Cross-Attention}, show that the Transformer-Encoder assigns high
attention scores to image regions containing rotation-informative image
cues, emphasizing vertical and horizontal lines. We also observe that the
proposed approach can predict the rotation of non-overlapping image pairs
with state-of-the-art (SOTA) accuracy. Our framework is end-to-end trainable
and optimizes a regression loss. It is evaluated on three dataset
benchmarks: StreetLearn ~\cite{mirowski2019streetlearn}, SUN360 \cite%
{xiao2012recognizing} and InteriorNet~\cite{InteriorNet18}, with different
overlap classes in indoor and outdoor locations and under varying
illumination. The experimental results in Section \ref{sec:experiment} show\
our model to provide state-of-the-art (SOTA) accuracy.

In summary, our contributions are as follows.

\begin{itemize}
\item We propose a novel scheme for estimating extreme rotations, including
scenarios with minimal image overlap.

\item Image embeddings are enhanced via cross-decoding, distilling
inter-image information.

\item A Transformer-Encoder cross-attention mechanism is proposed to encode
the latent space interactions between image pairs.

\item A decoder-decoder module infers relative rotation from the
cross-attention encoding by learning and applying quaternionic rotation
queries.

\item Quantitative evaluations demonstrate favorable performance compared to
state-of-the-art rotation estimation techniques on indoor and outdoor
datasets.
\end{itemize}

\section{Related Work}

Our rotation estimation approach represents a specific case of the more
general problem of relative pose estimation, particularly relative pose
regression (RPR). The prevalent relative rotation estimation technique
detects and matches 2D feature points [e.g., SIFT \cite{lowe2004distinctive},
SURF \cite{bay2006surf}] between images. For pose localization tasks \cite%
{taira2018inloc}, PnP schemes estimate the relative 3D rotation, and the
query image camera pose is determined given the anchor image's 3D
coordinates and pose. Other schemes for 3D rotation estimation utilized 3D Fourier transforms \cite{1467524,1710378}, whose magnitude is invariant to translations. Recent methods apply end-to-end trainable deep
networks to both images \cite{balntas2018relocnet,ding2019camnet}. Graph
neural networks (GNNs) enabled multi-image RPR via aggregating localization
cues across video frames \cite{9156582,9665967}. Neural radiance fields
(NeRFs) have been explored as an alternative to traditional image or feature
point storage for RPR encoding. Some schemes employ rotation-specific
parametrizations, notably quaternions and Euler angles, to estimate relative
3D rotations \cite{zhou2019continuity}. Such parametrizations, especially
quaternions, address the discontinuities intrinsic to rotation
representations, attributed to the Double-Cover property. Levinson et al. \cite{levinson2020analysis} investigated the SVD orthogonalization approach
for 3D rotation estimation via neural networks. By projecting the inferred
rotation matrices onto the rotation group using SVD, they showcased that its
integration supersedes conventional representations, advancing
the state-of-the-art in diverse deep learning paradigms. Further, Mohlin et al.~%
\cite{mohlin2020probabilistic} introduced a neural network-based estimation
of the parameters for the Fisher distribution matrix, representing the
probability distributions of 3D rotations. By optimizing the negative
log-likelihood loss of this distribution, they surpassed prior benchmarks in
several real-world datasets. Similarly, this optimization methodology was
used by Liu et al. to estimate the head pose \cite{9435939}. As a
noteworthy baseline, Rockwell et al.~\cite{8PointAlgorithm} devised a Vision
Transformer (ViT) to approximate the eight-point algorithm for direct
relative pose estimation between two images, showing competitive performance
across diverse scenarios. The methods above predominantly rely on
substantial overlap between input image pairs. A pronounced rotation,
resulting in limited overlap, could jeopardize the accuracy of these
estimations. Specifically, such techniques are ineffective for aligning
non-overlapping images. Caspi and Irani~\cite{caspi2002aligning}
demonstrated the feasibility of aligning two image sequences with
non-overlapping fields of view in both temporal and spatial dimensions,
provided the cameras are in proximity. This alignment leverages shared
temporal variations within the sequences. In a parallel vein, Shakil \cite%
{1660328} established that multiple nonoverlapping video sequences, captured
by uncalibrated video cameras, can be synchronized through inherent temporal
fluctuations and inter-frame motion within the sequences. Extending beyond
mere imagery, the challenge of registering non-overlapping RGB-D scans \cite%
{Sun_2022_CVPR,predator} serves as a notable derivative. In such cases, a
holistic representation of the scene is typically deduced.
In our study, we adopt the framework delineated by Cai et al. \cite%
{Cai2021Extreme}, where the task is to estimate the extreme relative 3D
rotation from a pair of input images. Cai et al. introduced a scheme in which a
CNN is used to embed the images, followed by the computation of a 4D
correlation volume (4DCV) from the embeddings. An MLP is then applied to
this correlation volume, optimizing it using a cross-entropy loss, resulting
in state-of-the-art (SOTA) accuracy for non-overlapping images.
Intrinsically, 4DCVs are an extension of Bilinear Pooling \cite%
{lin2015bilinear,8100226}, encoding pairwise correlations across all
entities of the 2D embedding maps corresponding to the image pair. Given
their encoding capability, 4DCVs have found applications in tasks
necessitating long-range spatial correspondences, evident in the RAFT SOTA
optical flow \cite{RAFT} and other optical flow models \cite%
{9495134,dosovitskiy2015flownet,NEURIPS2019_bbf94b34}. In a related context,
3D correlation volumes have been utilized in deep stereo matching tasks \cite%
{nie2019multi,liang2019stereo,gu2020cascade,9893813}, where the pixels in one image are matched with constrained spatial support in its counterpart.
Diverging from these methods, our proposal emphasizes the computation of an
analogous 4DCV by evaluating the cross-attention between the activation maps
of the image pair through multi-head Transformer-Encoder and an associated
activation mask. Specifically, this Transformer-Encoder effectively realizes
the functions of multiple aggregated correlation volumes \cite%
{NIPS2017_3f5ee243} via multi-head attention (MHA). Moreover, we propose an
inter-image embedding distillation using Transformer-Decoders, and also
improve the rotation inference using a cascaded alternating rotation
decoding.

\section{Rotation Estimation Using Cascaded Attention}

\label{sub:rotation estimation}

The proposed methodology estimates the relative 3D rotation $\mathbf{R}\in
\mathbb{R}^{3\times 3}$ between input image pairs ${\mathbf{I}_{1},\mathbf{I}%
_{2}} \in \mathbb{R}^{H\times W}$, outlined in Fig. \ref{fig:overview}. Siamese residual U-nets \cite%
{Zhang2018RoadEB} with weight sharing encode inputs into activation maps ${%
\widehat{\mathbf{I}}_{1},\widehat{\mathbf{I}}_{2}}\in \mathbb{R}^{c\times
K_{1}\times K_{2}}$, where $c$ is the number of channels and $K_{1}$, $K_{2}$
are spatial dimensions. To improve embeddings ${\widehat{\mathbf{I}}_{1},%
\widehat{\mathbf{I}}_{2}}$ via cross-decoding, ${\widehat{\mathbf{I}}_{1},%
\widehat{\mathbf{I}}_{2}}$ are cross-propagated into weight-sharing Transformer Decoder-0 units.
Each input embedding extracts task-relevant representations $\Bar{\Bar{%
\mathbf{I}}}_{1}$ and $\Bar{\Bar{\mathbf{I}}}_{2}$.

To further relate the two input images, we compute the cross-attention $%
\mathbf{\hat{T}}$, an enhanced equivalent of the 4D correlation volume
(4DCV) used in prior works \cite%
{RAFT,nie2019multi,liang2019stereo,9495134,dosovitskiy2015flownet}. Therefore, we have vectorized the rows $\Bar{\Bar{I_{1}}}$ and $\Bar{\Bar{I_{2}}}$ as two
sequences $\in \mathbb{R}^{c\times K_{1}K_{2}}$, concatenated as a single
tensor \textbf{$T$}$\in \mathbb{R}^{c\times 2K_{1}K_{2}}$, and apply a
Transformer-Encoder as in Section \ref{subsec:Cross-Attention}. The rotation
is decoded on the basis of cross-attention $\mathbf{\hat{T}}$ using a novel
attention-based architecture that uses a cascade of two
Transformer-Decoders. Initially, Transformer Decoder-1 is utilized to augment $\mathbf{\hat{T}}$ by incorporating it as a query, with guidance provided by a learnable quaternion $\mathbf{\Bar{q}} \in \mathbb{R}^{4}$ as input. Subsequently, the output of Transformer Decoder-1, denoted as $\mathbf{\Bar{\Bar{T}}}$, is introduced as input to the subsequent Transformer Decoder-2, further refining the query $\mathbf{\Bar{q}}$.  Finally, a fully connected
MLP layer is applied to predict the relative rotation encoded as quaternion $%
\Tilde{q}$. The 3D rotation is regressed using its quaternion
representation, $\mathbf{q,}$ as discussed in Section \ref{subsec:Regression}.

\subsection{Image Embedding Distillation by Cross-Decoding}

\label{subsec:cross-decoding}

Given the embeddings ${\widehat{\mathbf{I}}_{1},\widehat{\mathbf{I}}_{2}}\in
\mathbb{R}^{c\times K_{1}\times K_{2}}$ of the input images, we aim to
refine the embeddings by distilling the information between the input
images. For that, we apply a Transformer-Decoder that is applied to each
embedding, where the other embedding is used as the decoder's query. In
order to transform activation maps into Transformer-compatible inputs, we
follow the same sequence preparation procedure as in \cite{ObjectDetectionTransformers}. The
activation maps ${\widehat{\mathbf{I}}_{1},\widehat{\mathbf{I}}_{2}}\in
\mathbb{R}^{c\times K_{1}\times K_{2}}$ are first flattened to a sequential
representation ${\widehat{\mathbf{I}}_{1},\widehat{\mathbf{I}}_{2}}\in
\mathbb{R}^{c\times K_{1}K_{2}}$. Each position in the activation map is
further assigned with a learned encoding to preserve the spatial information
of each location. To reduce the number of parameters, two one-dimensional
encodings are learned separately for the $X$,$Y$ axes. Specifically, for an
activation map ${\widehat{\mathbf{I}}}$ we define the sets of positional
embedding vectors $\mathbf{E}_{u}\in \mathbb{R}^{K_{1}\times C/2}$ and $%
\mathbf{E}_{v}\in \mathbb{R}^{K_{2}\times C/2}$, such that a spatial
position $\left( i,j\right) ,$ $i\in 1..K_{1}$, $j\in 1..K_{2}$, is encoded
by concatenating the two corresponding embedding vectors:%
\begin{equation}
\mathbf{E}_{pos}^{i,j}=%
\begin{bmatrix}
\mathbf{E}_{u}^{i} \\
\mathbf{E}_{v}^{j}%
\end{bmatrix}%
\QQfntext{0}{{\scriptsize %
\url{https:/anonymous.4open.science/r/AttExtremeRotation-A467/}}}\in \mathbb{%
R}^{C}.
\end{equation}%
The processed sequence, serving as input to the Transformer is thus given
by:
\begin{equation}
\mathbf{\widehat{I}}={\widehat{\mathbf{I}}}+\mathbf{E_{A}}\in \mathbb{R}%
^{K_{1}K_{2}\times C},
\end{equation}%
where $\mathbf{E_{A}}$ is the positional encoding of ${\widehat{\mathbf{I}}}$%
.

\subsection{Cross-Attention Computation using a Transformer-Encoder}

\label{subsec:Cross-Attention}

The cross-attention between the refinement of representations of input
images $\Bar{\Bar{I_{1}}}$ and $\Bar{\Bar{I_{2}}}$ is computed using a
Transformer-Encoder with $l=2$ layers and $h=4$ attention heads for each
layer. An ablation study of this configuration is given in Section \ref%
{subsec:ablation}. The cross-attention maps computed by the
Transformer-Encoder are an improved equivalent of the 4D correlation volumes
\cite{RAFT,Cai2021Extreme}, encoding the interactions (inner-products)
between \textit{all} the image cues in the activation maps. By default, a
Transformer-Encoder computes the self-attention maps of the input sequence.
Hence, the cross-attention $\mathbf{\hat{T}}$ of the vectorized and
concatenated activation maps \textbf{$T$} is computed by applying the
attention mask $\mathbf{M}$ given in Eq. \ref{eq:mask}. The mask $\mathbf{M}$
nullifies the self-attention terms in the attention maps computed throughout
the Transformer-Encoder, while retaining the cross-attention terms,%
\begin{equation}
\mathbf{M}=%
\begin{bmatrix}
-\infty & -\infty & \cdots & 0 & 0 \\
-\infty & -\infty & \cdots & 0 & 0 \\
\vdots & \vdots & \ddots & \vdots &  \\
0 & 0 & \cdots & -\infty & -\infty \\
0 & 0 & \cdots & -\infty & -\infty%
\end{bmatrix}
\label{eq:mask}
\end{equation}

\begin{figure}[t]
\centering\includegraphics[width=\linewidth]{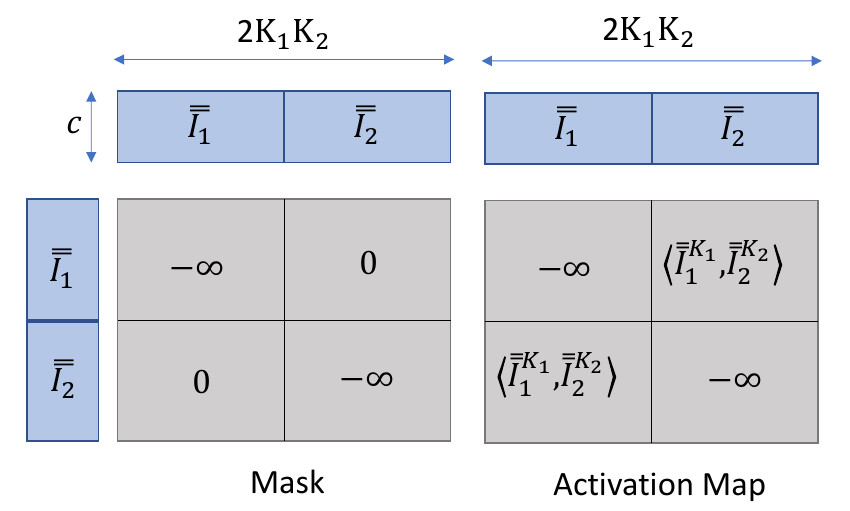} \vspace{-8pt}
\caption{Computing the cross-attention using a Transformer-Encoder and the
input mask, $\mathbf{M.}$ The mask $\mathbf{M}$ zeros the self-attention
terms, retaining only the cross-attention terms.}
\label{fig:mask}
\vspace{-8pt}
\end{figure}

The use of the mask $\mathbf{M}$ and the corresponding structure of the
attention maps is shown in Fig. \ref{fig:mask}.
Any pair of image patches could hold valuable information about the overall
geometric relationships in an image. The Transformer-Encoder can
uncover these hints implicitly. The regions in the input images that contain
rotation-related cues, explicitly or implicitly, receive higher attention
scores, as seen in Section 2 of the Supplementary Materials section. This
leads to a more meaningful and concise input for the distillation and the subsequent MLP layer,
ultimately improving the estimation accuracy. The same as in \cite%
{Cai2021Extreme}, even when the image pairs are non-overlapping, the
Transformer-Encoder formulation can predict the rotation using straight
lines only present in a single image, the same as human cognitive
capabilities. For example, in the extreme scenario of non-overlapping image
pairs the roll angle can be estimated from a single image, by implicitly
assuming that buildings and their edges are perpendicular to ground level.
Similarly, the relative elevation angle can be estimated by assuming that
the streets and pavements are parallel to the ground plane or by computing
the corresponding vanishing points. Most training and test datasets in this
domain depict urban scenes, adhering to these assumptions. \vspace{-8pt}

\subsection{Cascaded Attention-based Decoding}

\label{subsec:cascaded1} Given the cross-attention tensor $\mathbf{\hat{T}}$ that encodes interrelations between the paired input images, our objective is inferring the 3D relative rotation quaternion. To achieve this, we propose an innovative cascaded decoding scheme that alternately refines both the query rotation and cross-attention.
Initially, Transformer Decoder-1, enhances $%
\mathbf{\hat{T}}$ based on the learned quaternion $\overline{\mathbf{q}}$ to
compute $\overline{\overline{\mathbf{T}}}$. Next, the refined
cross-attention $\overline{\overline{\mathbf{T}}}$ is queried by $\overline{%
\mathbf{q}}$ to deduce the rotated encoding $\mathbf{\Bar{\Bar{q}}}$ using
Transformer Decoder-2. This cascaded inference approach could be extended
via additional dual units. However, we observed no performance gains from
additional cascades. Since the decoder inputs are semantic representations,
positional encodings are excluded.

\subsection{Relative Rotation Regression}

\label{subsec:Regression}

The encoded quaternion vector $\mathbf{\Bar{\Bar{q}}}$ is subsequently input
to a Multi-Layer Perceptron (MLP) regressor, computing the quaternion output
denoted $\mathbf{\Tilde{q}}$. This resultant quaternion is given by $\mathbf{%
q} = [q_w, q_x, q_y, q_z]$. The training loss is formulated as:
\begin{equation}
\mathbf{L}=| \ q_{0}-\mathbf{\Tilde{q}}/|| \mathbf{\Tilde{q}}|| \ |_2,
\label{equ:loss}
\end{equation}
where $q_{0}$ and $\Tilde{q}$ are the groundtruth and predicted quaternions,
respectively. Normalization ensures that the quaternion is a valid 3D
rotation representation.

\section{Experimental Results}

\label{sec:experiment}

The proposed scheme was experimentally verified by applying it to
contemporary benchmark datasets with overlapping and nonoverlapping image
pairs. Our experimental setup rigorously adhered to the paradigm established
by Cai et al. \cite{Cai2021Extreme}, using identical datasets and image
overlap categories. Utilizing their provided source code\footnote{%
{\scriptsize \url{https://github.com/RuojinCai/ExtremeRotation\_code}}}, to
create perspective views from panoramic images, ensuring that the input
images were the same in both studies, allowing fair comparisons with
previous SOTA results and other contemporary schemes. For that, we also used
the same Residual-Unet backbone network \cite{zhang2018road} as in \cite%
{Cai2021Extreme}. Section \ref{subsec:dataset} details the image datasets we
used and their processing, according to Cai et al. \cite{Cai2021Extreme}, to
derive the training and test datasets. Training details are given in Section %
\ref{subsec:Training Details}. We compare with recent SOTA schemes listed in
Section \ref{sec:baseline} using the geodesic error measure used in previous
work \cite{Cai2021Extreme}%
\begin{equation}
\mathbf{E}=\arccos \left( \frac{tr(\mathbf{R}^{T}\mathbf{R}^{\ast })-1}{2}%
\right) ,  \label{equ:GeodesicErr}
\end{equation}%
where $\mathbf{R}$ is the predicted rotation matrix and $\mathbf{R}^{\ast }$
is the groundtruth relative rotation matrix for each image pair. The
experimental comparisons are reported in Section \ref{subsec:comparisons}
and the attention maps are visualized in Section 2 of the Supplementary
Materials to provide an intuitive interpretation of the cross-attention
scores computed by the Transformer-Encoder. We studied the cross-dataset
generalization properties of the proposed scheme in Section \ref%
{subsec:multi-dataset}, while ablation studies of the different parameters,
design choices and parameters are reported in Section \ref{subsec:ablation}.

\subsection{Image Datasets and their Processing}

\label{subsec:dataset}

We used the following datasets and train/test splits used in previous works:

\textbf{InteriorNet}~\cite{InteriorNet18} is a synthetic data set to
understand and map interior scenes. A subset of 10,050 panoramas from 112
different houses was used, where the images of 82 houses were used for
training and those of 30 houses were used for testing, respectively.

\textbf{StreetLearn}~\cite{mirowski2019streetlearn} is an outdoor dataset
consisting of approximately 140,000 panoramic views of Pittsburgh and
Manhattan. We used 56K panoramic views from Manhattan, from which we
randomly chose 1000 panoramic views for testing.

\textbf{SUN360}~\cite{xiao2012recognizing} is an indoor collection of
high-resolution panoramas that cover a full view of $360^\circ \times
180^\circ$ for a variety of environmental scenes downloaded from the
Internet. It also provides location category labels. We used 7K and 2K
panoramas for training and testing, respectively.

As these datasets contain panoramic images, we generated 200 perspective $%
128\times 128$ images by randomly cropping 200 different locations in each
panoramic image. This sampling strategy ensures a consistent distribution of
ground-truth image pairs with pitch resolutions spanning $[-45^{\circ},
45^{\circ}]$ and yaw resolutions encompassing $[-180^{\circ}, 180^{\circ}]$.
We estimate only the Yaw and the Pitch angles, presuming a null roll between
paired images. We avoided generating textureless image pairs, that is,
images that mainly contain ceilings or floors in a house or skies in an outdoor
scenes, by limiting the pitch rang to $[-45^{\circ },45^{\circ }]$ for the
outdoor dataset and $[-30^{\circ },30^{\circ }]$ for the indoor datasets.
There is no overlap between the train and test datasets. To compare our
results with prior research and to analyze the influence of camera
translation on our rotation estimation approach, we partitioned the
InteriorNet and StreetLearn datasets into two groups: images with and
without camera translations. The non-translated images were acquired by
randomly selecting pairs of cropped images from a single panorama. In
contrast, datasets that include translations (known as StreetLearn-T and
InteriorNet-T) were generated by randomly selecting pairs of cropped images
from different panoramas, where translations are less than $3m$. However,
our method was not used to estimate these translations. We evaluated our
performance in overlapping and nonoverlapping pairs and use the setup of Cai
et al.~\cite{Cai2021Extreme} by dividing the datasets into three overlap
classes:

\textbf{Large}, contains highly overlapping pairs up to relative rotations
of $45^\circ$

\textbf{Small}, contains pairs that partially overlap with relative rotation
angles $\in \lbrack 45^\circ,90^\circ]$

\textbf{None}, contains pairs without overlap with relative rotations $%
>90^\circ$.

\subsection{Training Details}

\label{subsec:Training Details}

We use a pre-trained Residual-Unet \cite{zhang2018road} (same as in Cai et
al. \cite{Cai2021Extreme}) as a backbone to compute the feature maps of the
two input images \textbf{$(\hat{I_{1}},\hat{I_{2}})$}$\in \mathbb{R}%
^{c\times k_{1}\times k_{2}}=\mathbb{R}$$^{128\times 32\times 32}$.
According to Fig. \ref{fig:overview}, subsequently, these feature maps are
cross-propagated into dedicated decoder units, resulting in the refinement
of representations $\Bar{\Bar{I_{1}}}$ and $\Bar{\Bar{I_{2}}}$. The
refinements of the representations $\Bar{\Bar{I_{1}}}$ and $\Bar{\Bar{I_{2}}}
$ were reshaped and concatenated along the axis of the samples to form the
tensor $\mathbf{T}\in \mathbb{R}^{(2\cdot 32\cdot 32)\times 128}=\mathbb{R}%
^{2048\times 128}$. $\mathbf{T}$ was the input to the Transformer-Encoder,
consisting of $l=2$ layers with ReLU nonlinearity and a dropout of $p=0.1$.
Each encoder layer uses $h=4$ MHA heads and a hidden dimension of $C_{h}=768$%
. An ablation study of the Transformer-Encoder parameters is given in
Section \ref{subsec:ablation}. The Transformer-Encoder's output \textbf{$%
\hat{T}$} is then fed into a dual-path structure comprising two concatenated
decoders. The primary decoder receives a learnt quaternion vector
initialized by white Gaussian noise, $\Bar{q}$, as input and the
cross-attention $\hat{T}$ as a query, and produces $\Bar{\Bar{T}}$,
enhancing contextual nuances, while the subsequent decoder gets $\Bar{\Bar{T}%
}$ as an input and the same empty quaternion vector, $\Bar{q}$ as a query,
and generates $\Bar{\Bar{q}}$, encapsulating pivotal rotational attributes.
The two sequential attention-based decoders use $l=2$ layers with $h=2$ MHA
heads and a hidden dimension of $C_{h}=768$. Finally, the MLP regressor that
computes the quaternion representation for the regression loss in Eq. \ref%
{equ:loss}. The MLP regressor contains two fully connected layers.
Throughout all experiments, the model is optimized using an Adam optimizer
with an initial learning rate of $\lambda =5e-4$, with $\beta _{1}=0.9$, $%
\beta _{2}=0.999$, $\epsilon =10^{-10}$\textbf{, }and a batch size of $20$.
Our model is implemented in PyTorch, it is end-to-end trainable, and all
experiments were performed on an 8GB NVIDIA GeForce GTX 2080 GPU.%
\definecolor{graytext}{RGB}{130,130,130}

\begin{table*}[t]
\setlength{\tabcolsep}{1.0pt}
 \def\arraystretch{0.95}
\centering
\resizebox{\textwidth}{!}{%
\begin{tabular}{llccclccclccclccclccc}
\toprule 
\multicolumn{2}{l}{}& \multicolumn{3}{c}{InteriorNet} &  & \multicolumn{3}{c}{InteriorNet-T} &  & \multicolumn{3}{c}{SUN360} &  & \multicolumn{3}{c}{StreetLearn} &  & \multicolumn{3}{c}{StreetLearn-T}\\ \cline{3-5} \cline{7-9} \cline{11-13} \cline{15-17} \cline{19-21}
Overlap
& Method
& Avg(\degree$\downarrow$) & Med(\degree$\downarrow$) & \multicolumn{1}{c}{10\degree(\%$\uparrow$)} & & Avg(\degree$\downarrow$) & Med(\degree$\downarrow$) & \multicolumn{1}{c}{10\degree(\%$\uparrow$)} &  & Avg(\degree$\downarrow$) & Med(\degree$\downarrow$) & \multicolumn{1}{c}{10\degree(\%$\uparrow$)} &  & Avg(\degree$\downarrow$) & Med(\degree$\downarrow$) & \multicolumn{1}{c}{10\degree(\%$\uparrow$)} &  & Avg(\degree$\downarrow$) & Med(\degree$\downarrow$) & \multicolumn{1}{c}{10\degree(\%$\uparrow$)} \\ \midrule 
\multirow{7}{*}{Large} & SIFT*~\cite{lowe2004distinctive}       & 6.09                 & 4.00                 & 84.86               &  & 7.78                 & 2.95                 & 55.52               &  & 5.46                       & 3.88                       & 93.10                      &  & 5.84                 & 3.16                 & 91.18               &  & \color{graytext}{18.86}                & \color{graytext}{3.13}                 & \color{graytext}{22.37}               \\
                         & SuperPoint*~\cite{detone18superpoint}  & 5.40                 & 3.53                 & 87.10               &  & 5.46                 & 2.79                 & 65.97               &  & 4.69                       & 3.18                       & 92.12                      &  & 6.23                 & 3.61                 & 91.18               &  & \color{graytext}{6.38}        & \color{graytext}{1.79}        & \color{graytext}{16.45}               \\
                         & Reg6D~\cite{zhou2019continuity}        & 9.05                 & 5.90                 & 68.49               &  & 17.00                & 11.95                & 41.79               &  & 16.51                      & 12.43                      & 40.39                      &  & 11.70                & 8.87                 & 58.24               &  & 36.71                & 24.79                & 23.03               \\
                         & DenseCorrVo ~\cite{Cai2021Extreme}                                & 1.53        & 1.10                 & 99.26     &  & 2.89        & 1.10        & 97.61     &  & 1.00              & 0.94              & \textbf{100.00}           &  & 1.19        & 1.02        & 99.41   &  & 9.12                 & 2.91                 & 87.50     \\
                         & 8PointViT ~\cite{8PointAlgorithm}                                   & 0.48                 & 0.40        & \textbf{100.00}               &  & 2.90                 & 1.83                 & 97.91 &  &   - &  -  &  -  & & 0.62                       & 0.52                       & \textbf{100.00}                      &  & 4.08                 & 2.43                 & \textbf{90.13}         \\
                         & Ours                                 & \textbf{0.43}                  & \textbf{0.38}                 &  99.65              &  & \textbf{1.75}                 & \textbf{0.95}                 & \textbf{98.8}               &  & \textbf{0.85}                        & \textbf{0.45}                      & 99.95                      &  & \textbf{0.58}                 & \textbf{0.48}                         & 99.31       &  & \textbf{3.88}                  &  \textbf{1.69}                        & 87.20  \\ \midrule 
\multirow{6}{*}{Small} & SIFT*~\cite{lowe2004distinctive}       & 24.18                & 8.57                 & 39.73               &  & \color{graytext}{18.16}                & \color{graytext}{10.01}                & \color{graytext}{18.52}               &  & 13.71                      & 6.33                       & 56.77                      &  & 16.22                & 7.35                 & 55.81               &  & \color{graytext}{38.78}                & \color{graytext}{13.81}                & \color{graytext}{5.68}                \\
                         & SuperPoint*~\cite{detone18superpoint}  & \color{graytext}{16.72}                & \color{graytext}{8.43}                 & \color{graytext}{21.58}               &  & \color{graytext}{11.61}                & \color{graytext}{5.82}                 & \color{graytext}{11.73}               &  & \color{graytext}{17.63}                      & \color{graytext}{7.70}                       & \color{graytext}{26.69}                      &  & \color{graytext}{19.29}                & \color{graytext}{7.60}                 & \color{graytext}{24.58}               &  & \color{graytext}{6.80}        & \color{graytext}{6.85}                 & \color{graytext}{0.95}                \\
                         & Reg6D~\cite{zhou2019continuity}        & 25.71                & 15.56                & 33.56               &  & 42.93                & 28.92                & 23.15               &  & 42.55                      & 32.11                      & 9.40                       &  & 24.77                & 15.11                & 30.56               &  & 46.61                & 34.33                & 13.88               \\
                         & DenseCorrVol ~\cite{Cai2021Extreme}                                 & 6.45                 & 1.61                 & 95.89               &  & 10.24      & 1.38        & 89.81    &  & 3.09              & 1.41              & 98.50           &  & 2.32       & 1.41       & 98.67     &  & 13.04                & 3.49                 & 84.23     \\
                          & 8PointViT ~\cite{8PointAlgorithm}                                   & 1.84              & 0.94        & 99.32               &  & 4.48                 & 2.38                 & 96.30 &  &   - &  -  &  -  & & 1.46                       & 1.09                       & \textbf{100.00}                      &  & 9.19                 & 3.25                 & 87.7         \\                   
                         & Ours                             & \textbf{1.55}                 & \textbf{0.872}                 & \textbf{99.85}               &  & \textbf{4.25}      & \textbf{0.777}        &  \textbf{97.55}     &  & \textbf{2.109}              & \textbf{0.831}              & \textbf{98.99}            &  & \textbf{1.21}        & \textbf{0.718}        & 99.122    &  & \textbf{7.48}                & \textbf{1.8666}                 & \textbf{88.996}     \\   \midrule       
\multirow{5}{*}{None}   & SIFT*~\cite{lowe2004distinctive}       & \color{graytext}{109.30}               & \color{graytext}{92.86}                & \color{graytext}{0.00}                &  & \color{graytext}{93.79}                & \color{graytext}{113.86}               & \color{graytext}{0.00}                &  & \color{graytext}{127.61}                     & \color{graytext}{129.07}                     & \color{graytext}{0.00}                       &  & \color{graytext}{83.49}                & \color{graytext}{90.00}                & \color{graytext}{0.38}                &  & \color{graytext}{85.90}                & \color{graytext}{106.84}               & \color{graytext}{0.38}                \\
                         & SuperPoint*~\cite{detone18superpoint}  & \color{graytext}{120.28}               & \color{graytext}{120.28}               & \color{graytext}{0.00}                &  & \color{graytext}{--}                   & \color{graytext}{--}                   & \color{graytext}{0.00}                &  & \color{graytext}{149.80}                     & \color{graytext}{165.24}                     & \color{graytext}{0.00}                       &  & \color{graytext}{--}                   & \color{graytext}{--}                   & \color{graytext}{0.00}                &  & \color{graytext}{--}                   & \color{graytext}{--}                   & \color{graytext}{0.00}                \\                   
                         & Reg6D~\cite{zhou2019continuity}        & 48.36                & 32.93                & 10.82               &  & 60.91                & 51.26                & 11.14               &  & 64.74                      & 56.55                      & 3.77                       &  & 28.48                & 18.86                & 24.39               &  & 49.23                & 35.66                & 11.86               \\
                         & 8PointViT ~\cite{8PointAlgorithm}                                   & -              & -       & -               &  & -                 & -                 & -&  &   - &  -  &  -  & & -                       & -                       & -                      &  & -                & -                & -         \\
                         & DenseCorrVol ~\cite{Cai2021Extreme}                                   & 37.69       & 3.15       & 61.97     &  & 49.44       & 4.17        & 58.36     &  & 34.92             & 4.43             & 61.39           &  & 5.77       & 1.53        & \textbf{96.41}     &  & 30.98       & 3.50       & \textbf{72.69}   \\
                          & Ours                                & \textbf{35.13}       & \textbf{2.814}        & \textbf{65.20}     &  & \textbf{45.32}       & \textbf{4.05}       & \textbf{59.56}     &  & \textbf{32.46}            & \textbf{4.19}             & \textbf{63.17}            &  & \textbf{5.33}        & \textbf{1.20}        & 96.22     &  & \textbf{28.13}       &\textbf{ 3.25}        & 72.43     \\ \midrule
\end{tabular}
}
\vspace{-8pt}
\caption{Relative rotation estimation results. We utilized the InteriorNet, SUN360, and StreetLearn datasets and show the average and median of geodesic errors. We also present the percentage of image pairs with relative rotation error below 10$\degree$, for the overlap categories in Section \ref{subsec:dataset}. The gray numbers indicate errors exceeding $50\%$. The asterisk $*$ signifies that the mean and median errors did not lead to pose estimation, and their calculations are performed only on successful image pairs.}
\label{tab:main_result}
\end{table*}

\subsection{Comparative baselines}

\label{sec:baseline}

In line with Cai et al. \cite{Cai2021Extreme}, we compare our method with
contemporary schemes using the datasets in Section \ref{subsec:dataset}:

\textbf{A} \textbf{SIFT-based approach \cite{brown2007minimal}}. A method
for matching SIFT features \cite{lowe2004distinctive} using RANSAC \cite%
{fischler1981random} in image pairs of the same panorama, and estimating the
relative rotation matrix using Homography equations or the Essential matrix.

\textbf{CNN-based methods \cite{detone18superpoint}}. Deep learning schemes
that detect and encode local image features using SuperPointNet~\cite%
{detone18superpoint} and D2-Net~\cite{Dusmanu2019CVPR}.

\textbf{Self-supervised interest point \cite{zhou2019continuity}. }A scheme
by Zhou et al. \cite{zhou2019continuity} (Reg6D) that applies a CNN to
approximate the mappings between various rotation representations and fits
continuous 5D and 6D rotation representations, instead of the commonly used
Euler and quaternion representations.

\textbf{Extreme rotation estimation \cite{Cai2021Extreme}}. A deep learning
technique to estimate the relative 3D rotation of image pairs in an extreme
setting \cite{Cai2021Extreme} where the images have little or no overlap.
They proposed a network that automatically learns implicit visual cues by
computing a 4D correlation volume.

\textbf{Attention-based methods.} We also compare to recent work by Rockwell
et al. \cite{8PointAlgorithm} (8PointVit) using a Vision Transformer (ViT)
to estimate the relative pose. Although Rockwell et al. achieve competitive
results in multiple settings, their approach is less suited for extreme view
changes.

\subsection{Experimental comparisons}

\label{subsec:comparisons}

The results of the comparison of the proposed scheme with baselines and SOTA
schemes are reported in Table~\ref{tab:main_result}. We report the mean and
median of the geodesic error given in Eq. \ref{equ:GeodesicErr} and the
percentage of image pairs whose estimated relative rotation error was less
than ${10^{\circ }}$. We compared the accuracy of our proposed model to the
schemes detailed in Section \ref{sec:baseline}. The proposed approach is
shown to be accurate for both indoor and outdoor scenes and significantly
outperforms the baseline schemes in all overlap categories. For
nonoverlapping pairs, correspondence-based methods such as SIFT \cite%
{brown2007minimal}, SuperPointNet~\cite{detone18superpoint}, Reg6D~\cite%
{zhou2019continuity} and 8PointViT \cite{8PointAlgorithm} failed to provide
any estimates, as they require feature correspondence. The
DenseCorrVol~approach \cite{Cai2021Extreme} provides accurate results in
extreme cases, but our approach outperforms it. Qualitative experimental
results are given in Section 1 of the Supplementary Materials section.
\begin{figure*}[tbh]
\centering
\begin{tabular}{c|cc|cc}
\raisebox{2em}{\rotatebox[origin=c]{90}{Large}} & %
\includegraphics[width=0.22\linewidth]{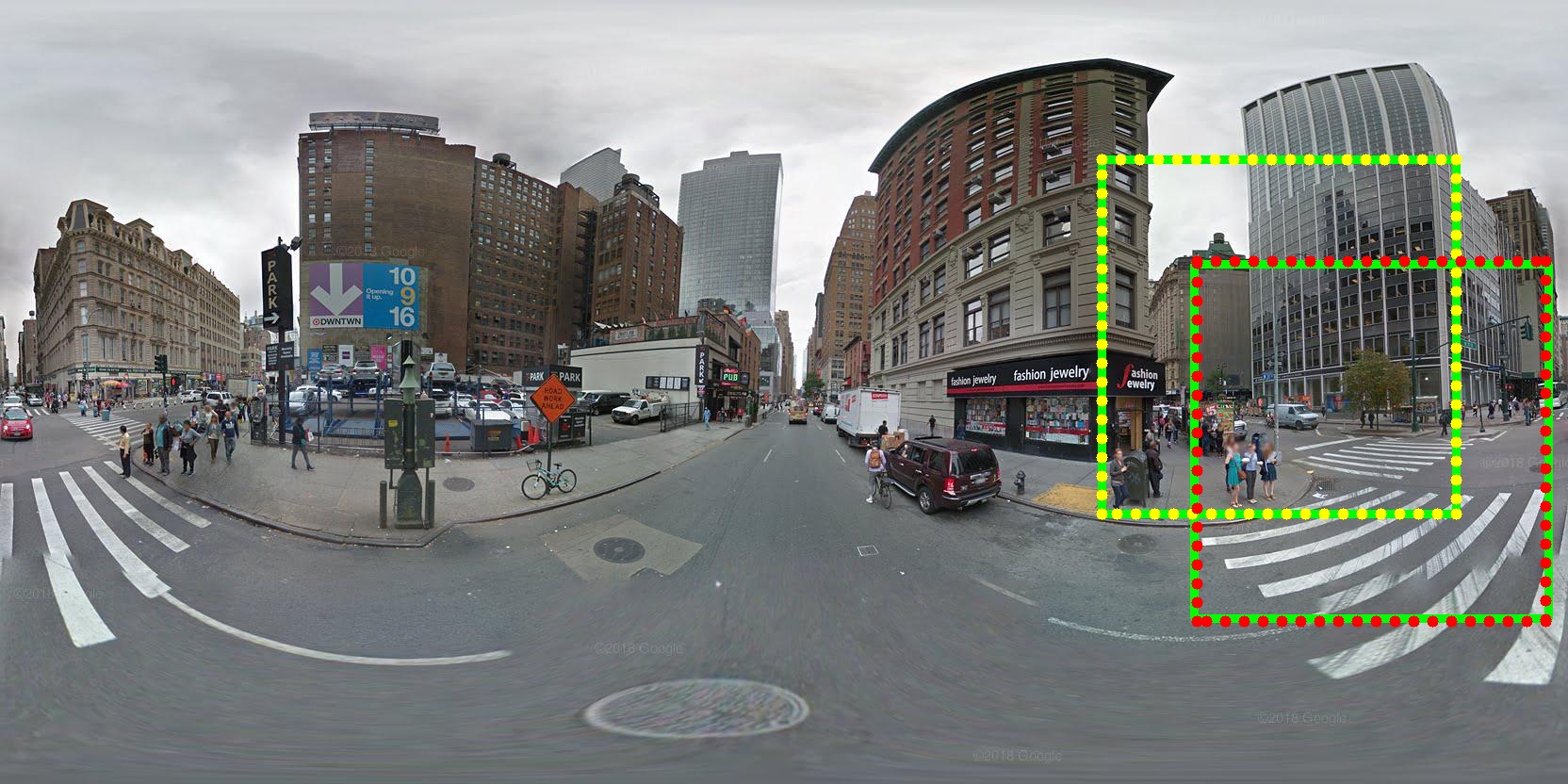} & %
\includegraphics[width=0.22\linewidth]{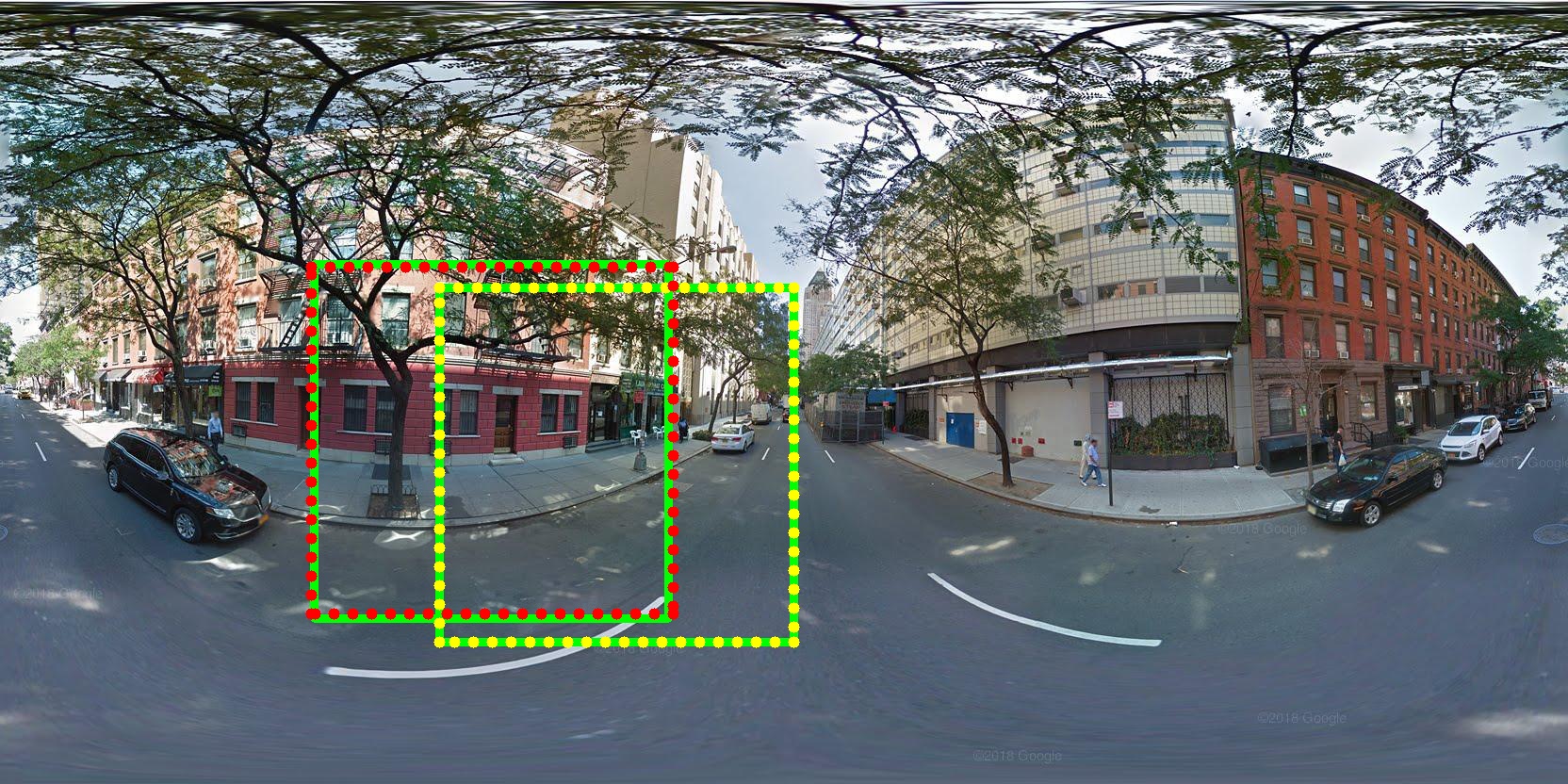} & %
\includegraphics[width=0.22\linewidth]{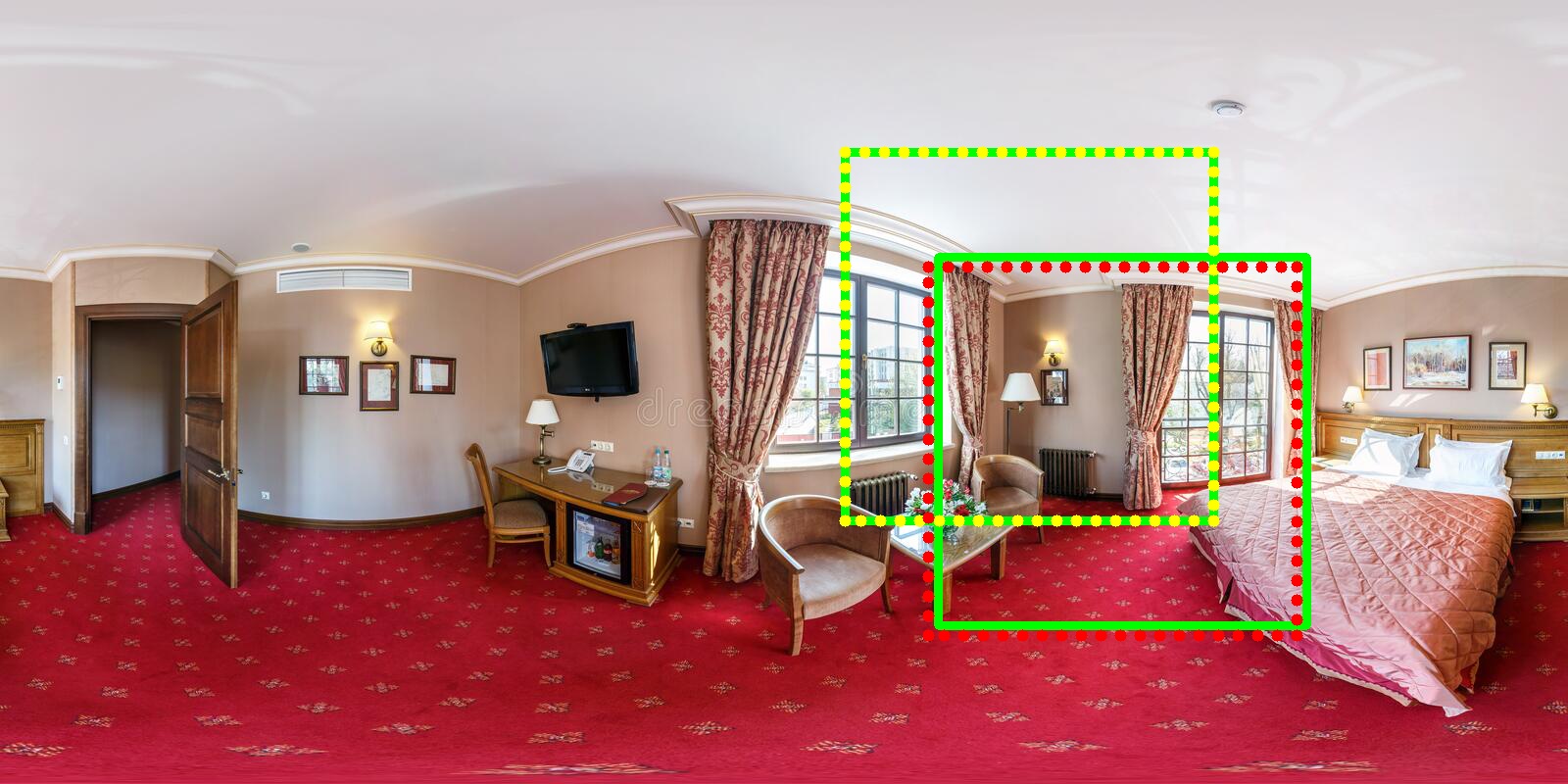} & %
\includegraphics[width=0.22\linewidth]{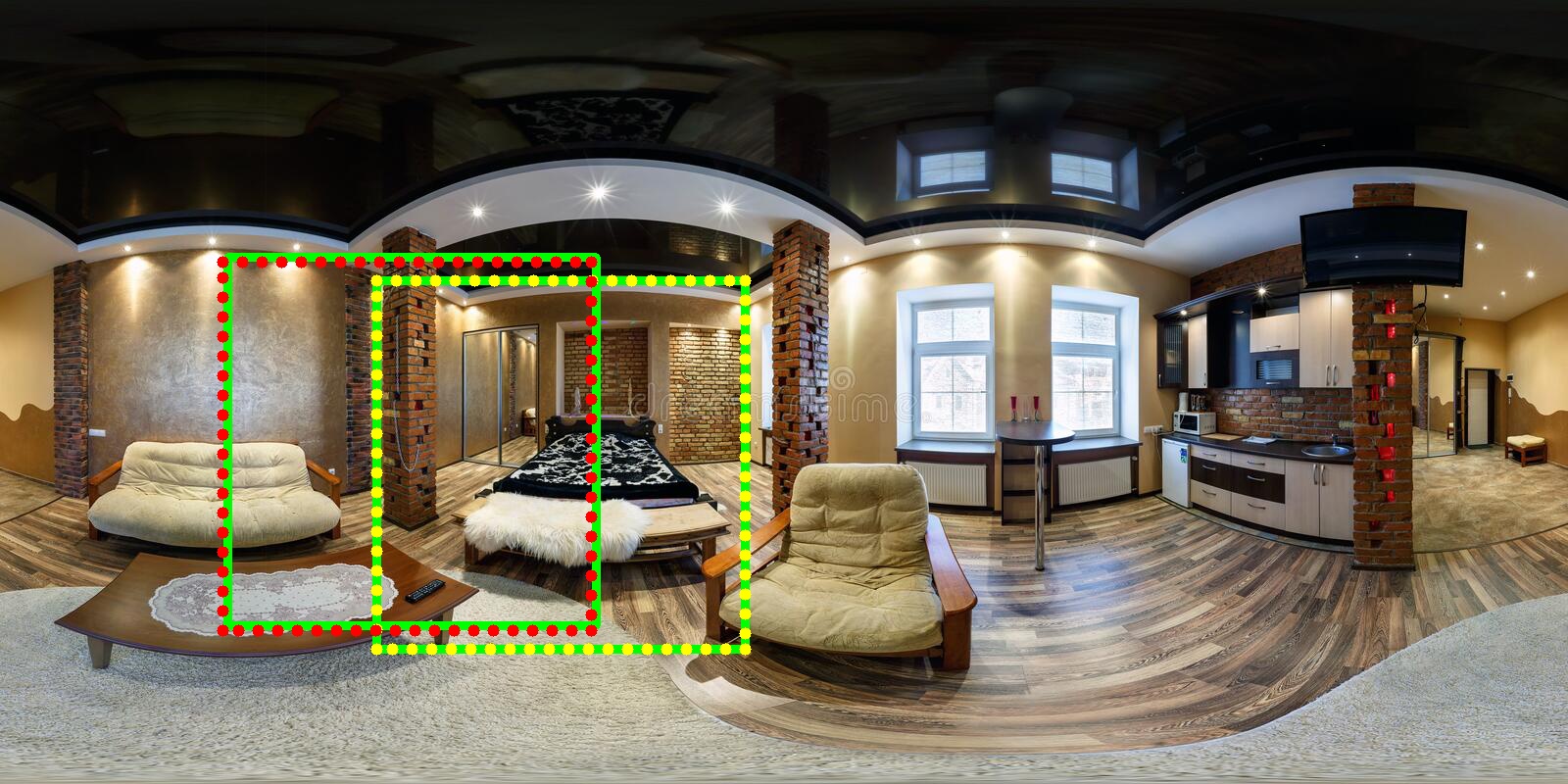} \\
\raisebox{2em}{\rotatebox[origin=c]{90}{Small}} & {%
\includegraphics[width=0.22\linewidth]{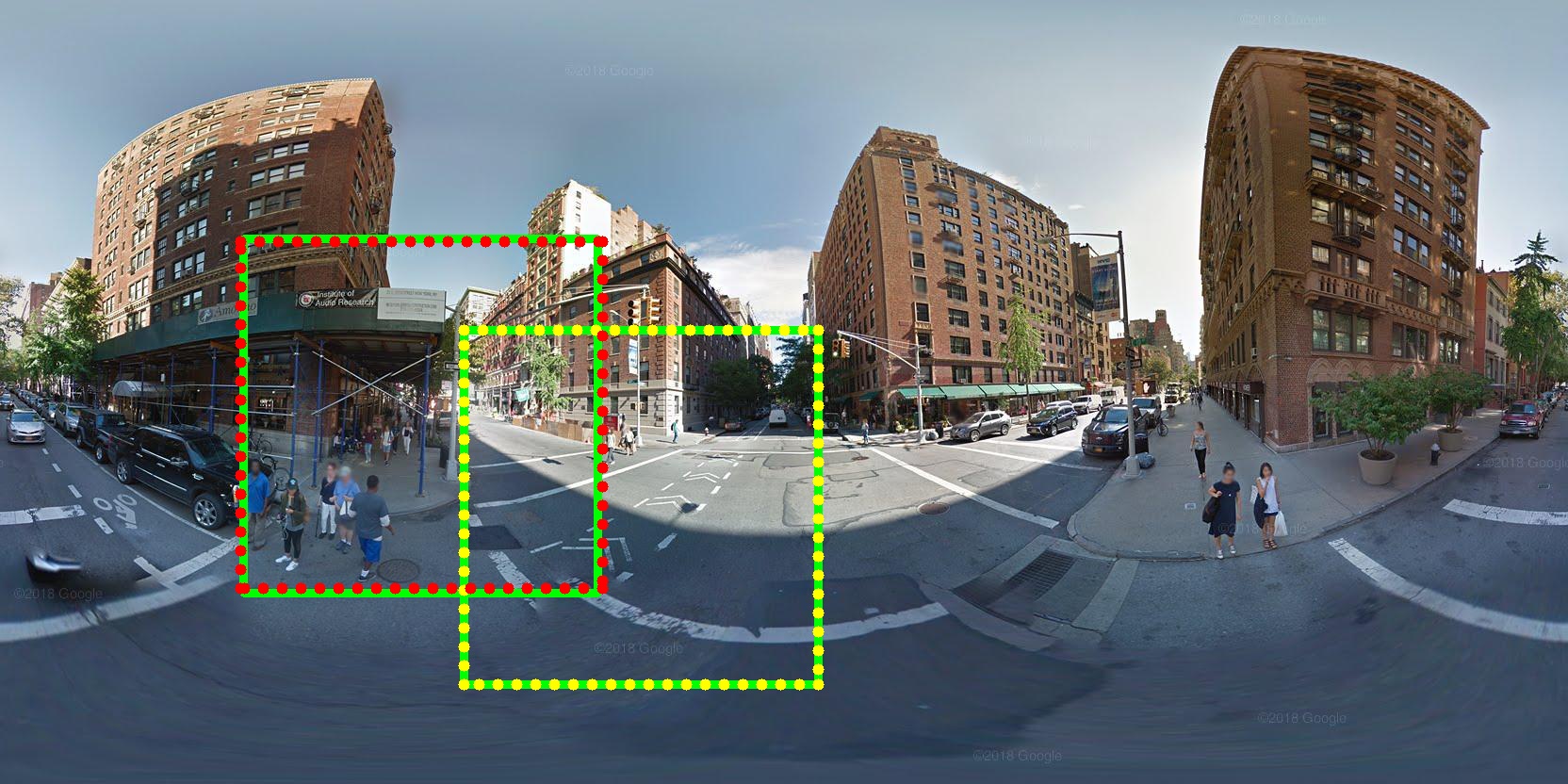}} & {%
\includegraphics[width=0.22\linewidth]{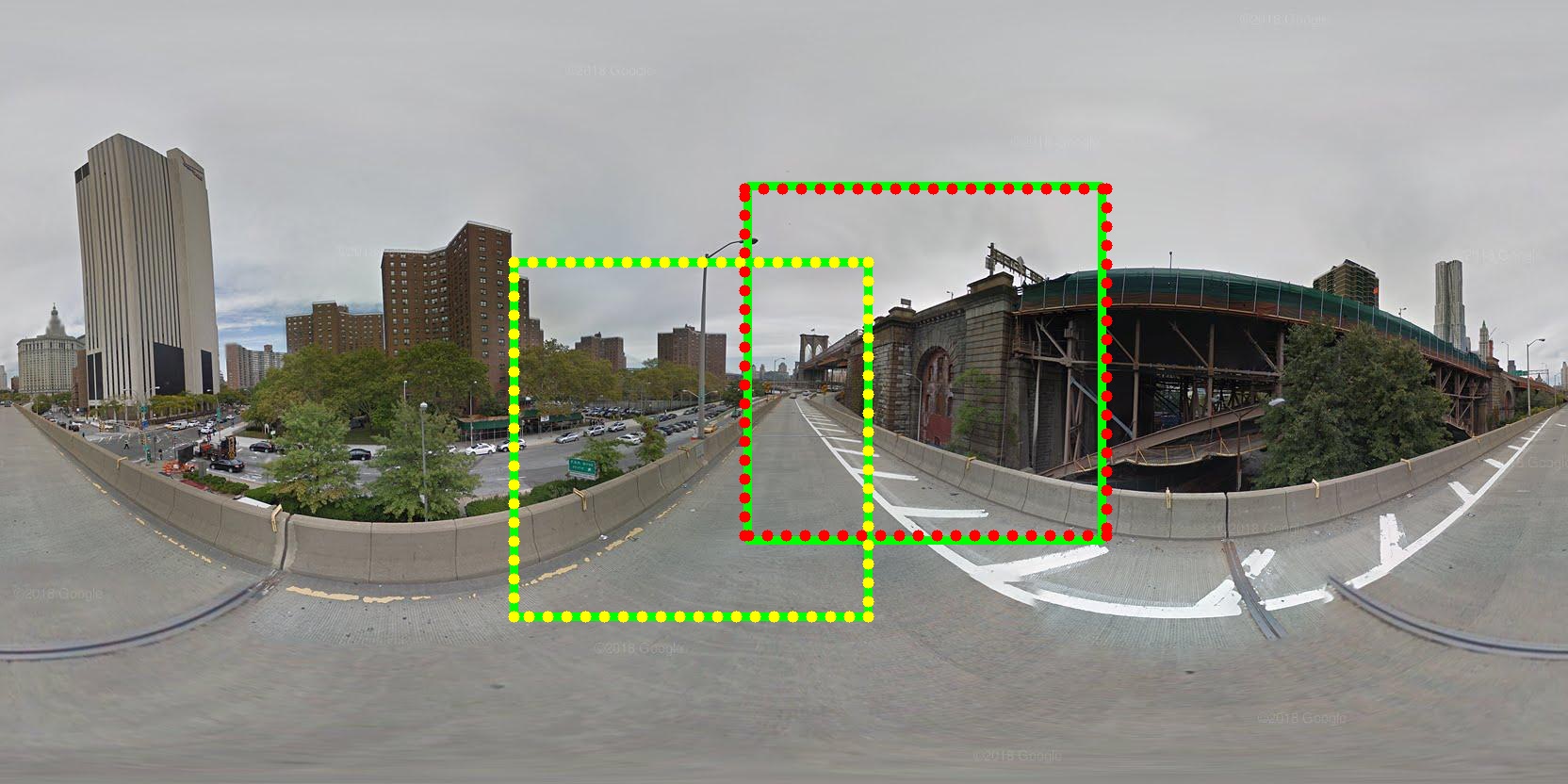}} & %
\includegraphics[width=0.22\linewidth]{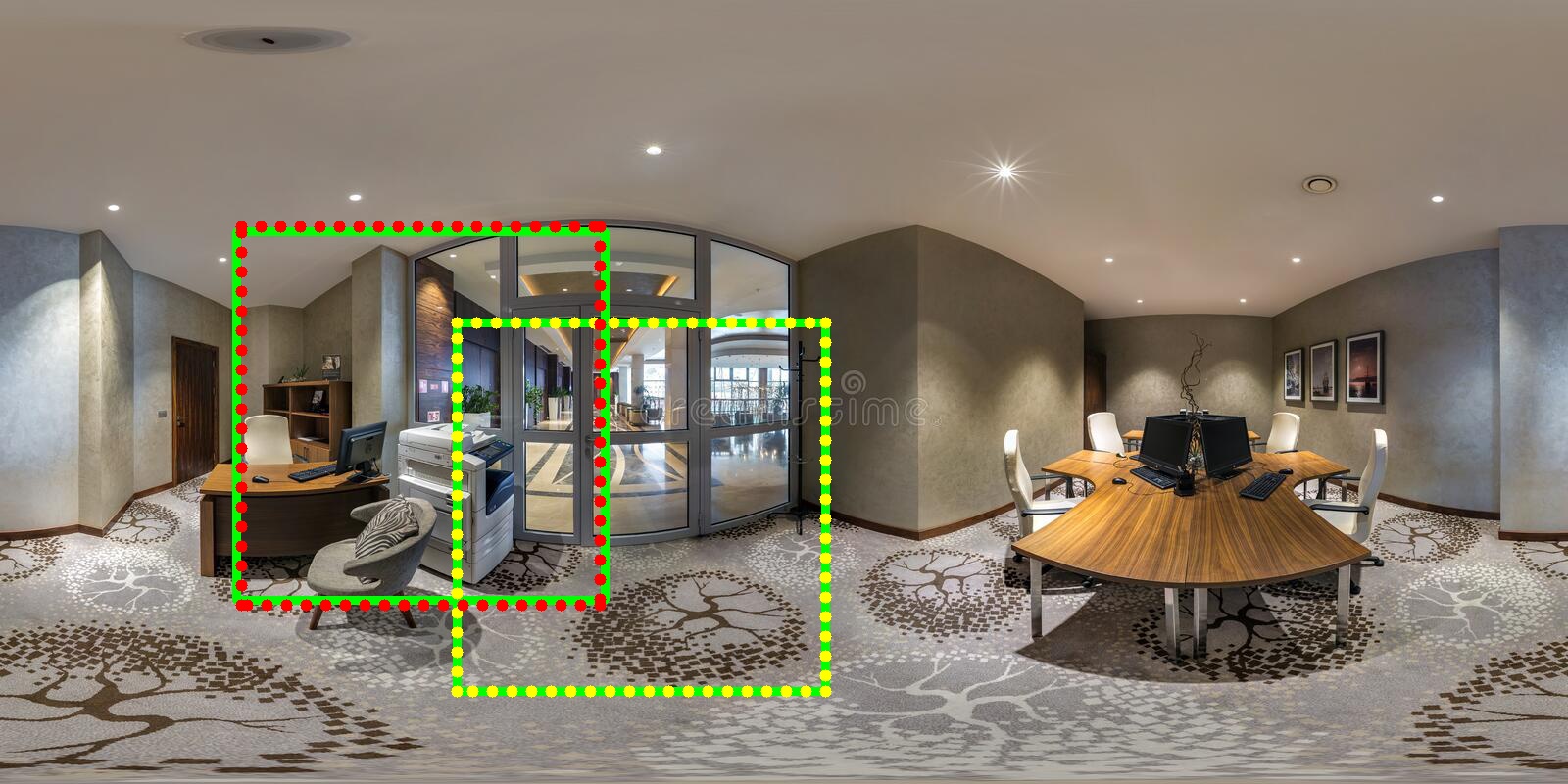} & {%
\includegraphics[width=0.22\linewidth]{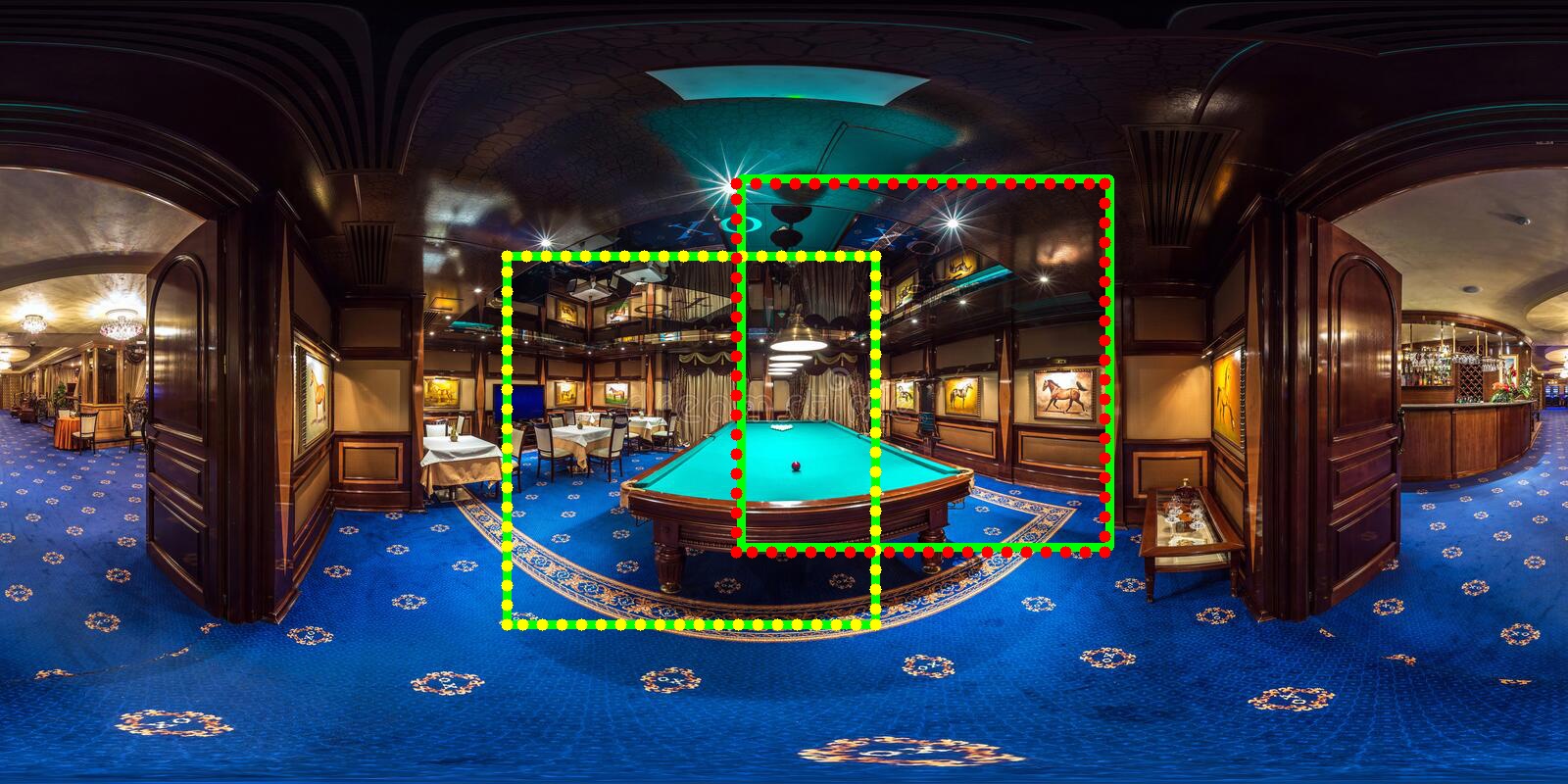}} \\
\raisebox{2em}{\rotatebox[origin=c]{90}{No}} & {\includegraphics[width=0.21%
\linewidth]{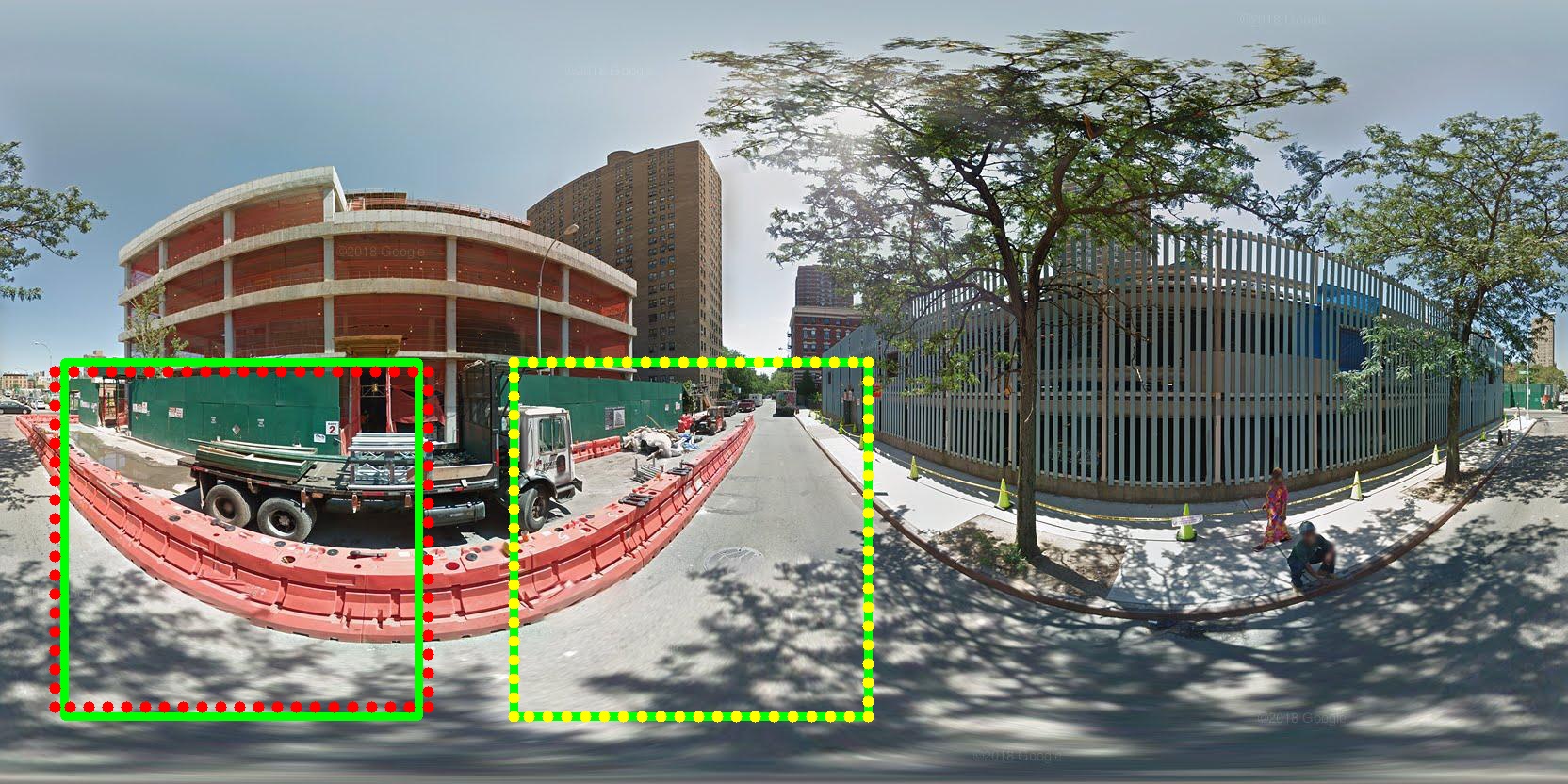}} & {\includegraphics[width=0.22%
\linewidth]{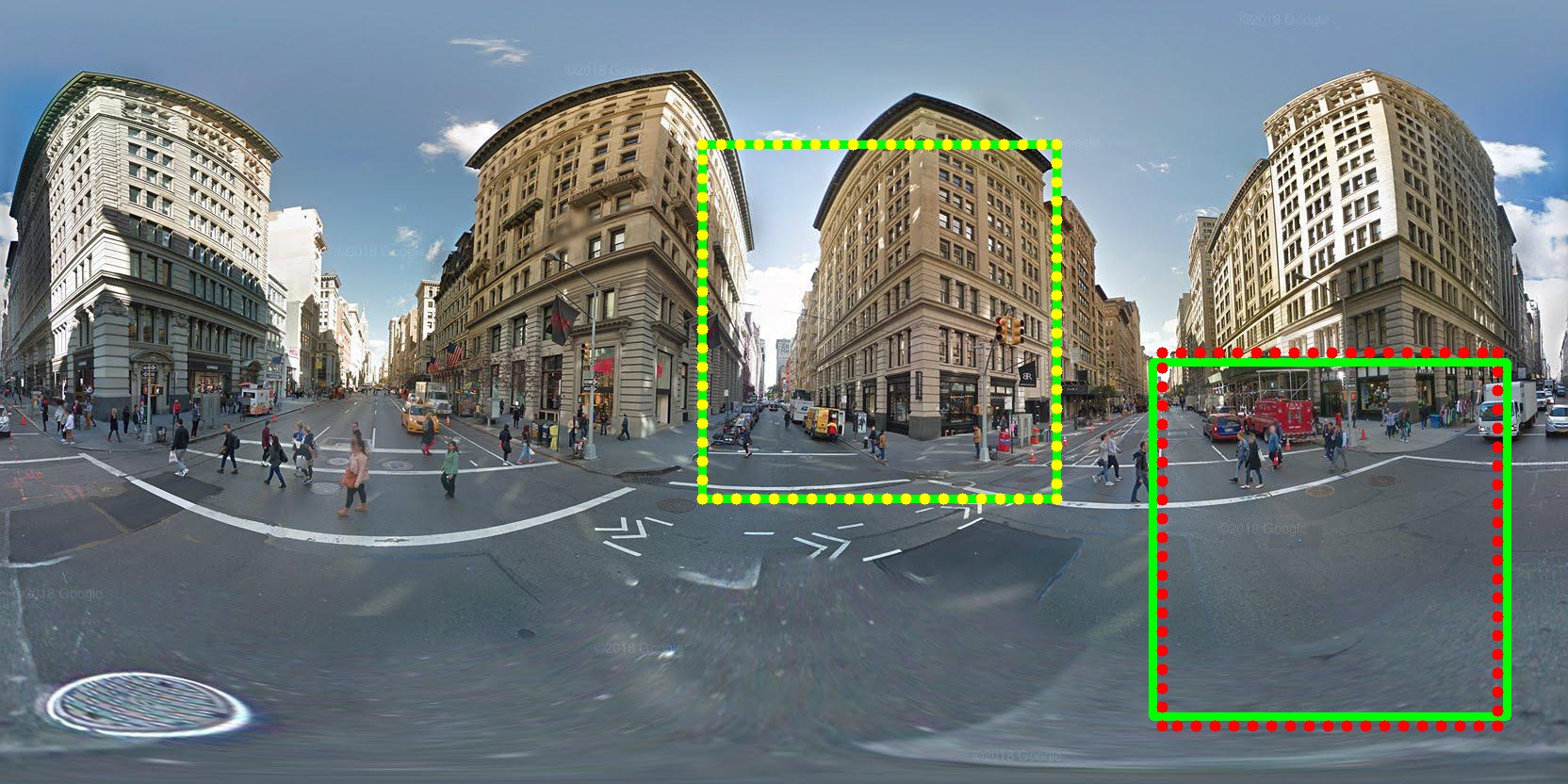}} & \includegraphics[width=0.22%
\linewidth]{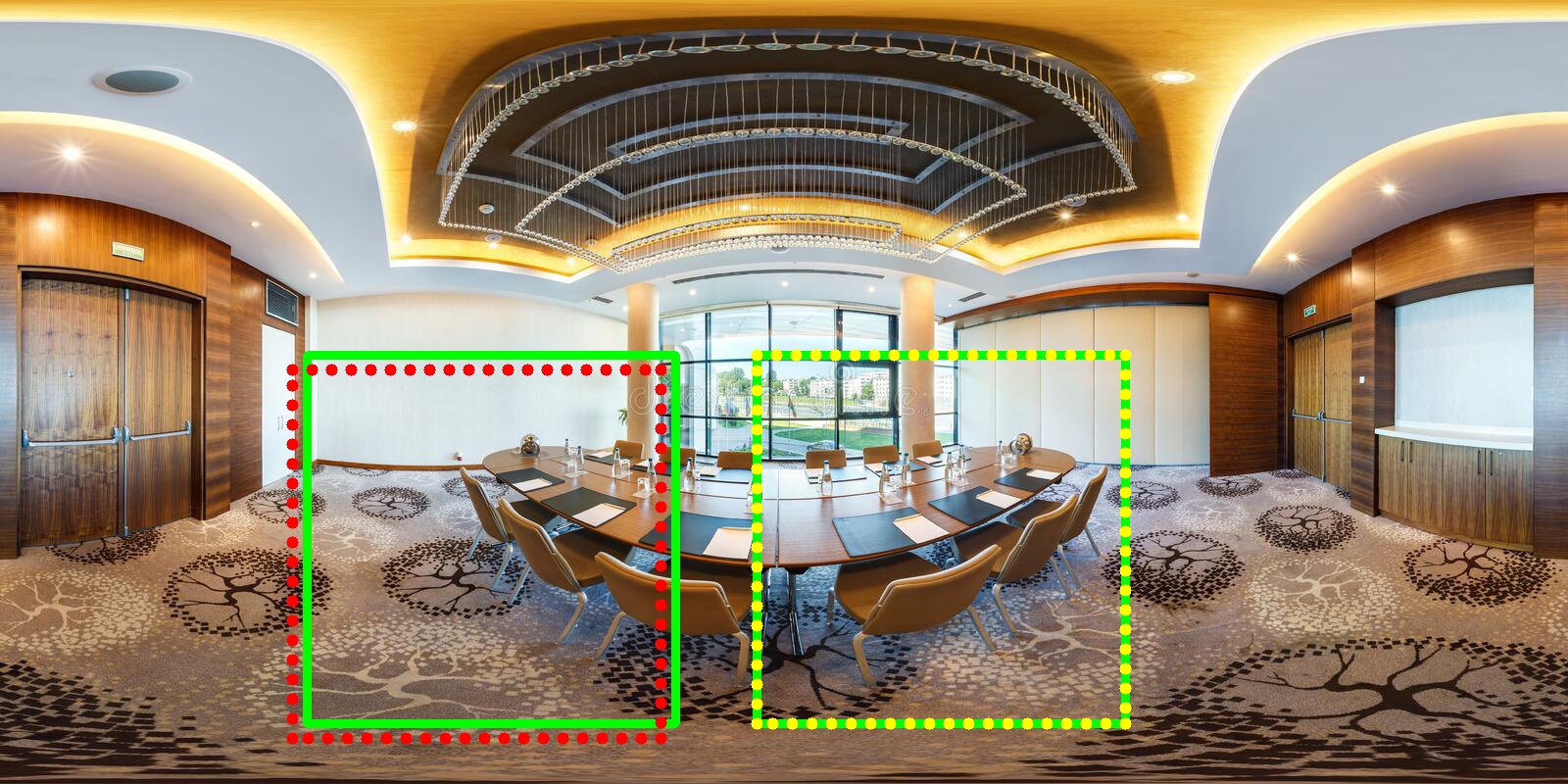} & {\includegraphics[width=0.22%
\linewidth]{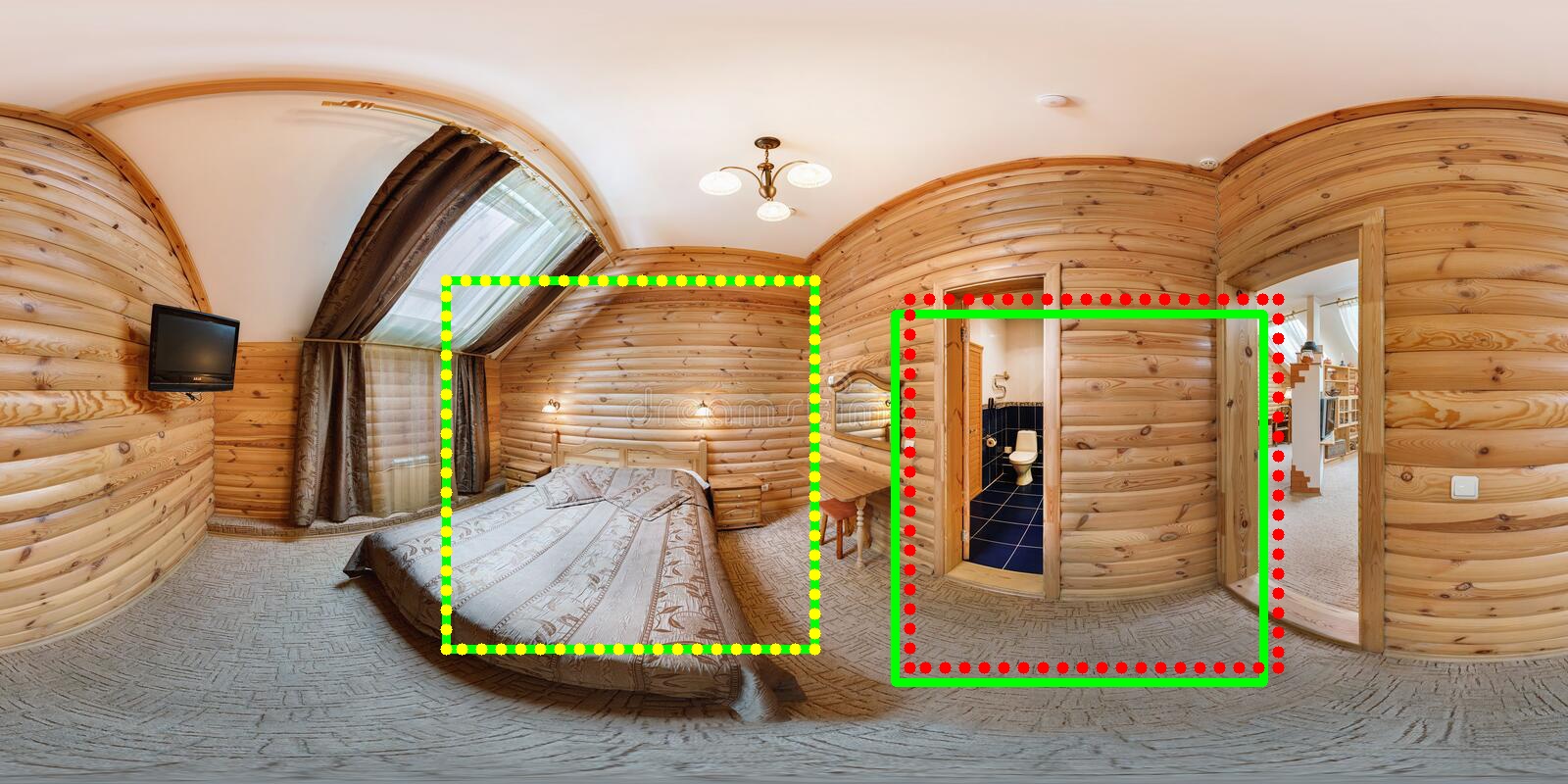}} \\
& \multicolumn{2}{c|}{StreetLearn} & \multicolumn{2}{c}{SUN360}%
\end{tabular}%
\vspace{-8pt}
\caption{Rotation estimation results. The panoramic and cropped groundtruth
images are marked by green and yellow-dot lines. The predicted footprint of
one of the cropped images is marked by the red-dot line. The first row shows
the matching results of images with large overlaps. The second and last rows
show the matching of small overlap and non-overlapping images.}
\label{fig:predicted}
\end{figure*}
The qualitative results of the rotation estimation are shown in Fig. \ref%
{fig:predicted} for the StreetLearn and SUN360 datasets, for the large,
small and nonoverlapping cases. We show the full panoramas, the footprints
of the cropped images that were used as inputs for the proposed scheme and
the footprint of the estimated image crop based on the estimated rotation.
In all cases, we archieve high estimation accuracy.

\subsection{Cross-Dataset Generalization}
\label{subsec:multi-dataset}
\begin{table}[tbh]
\centering
\begin{tabular}{lcc}
\toprule Overlap & Ours [\degree] & Cai et al. \cite{Cai2021Extreme} [\degree%
] \\
\midrule Large & \textbf{9.55} & 11.23 \\
Small & \textbf{16.33} & 20.87 \\
None & \textbf{38.48} & 40.82 \\
\bottomrule &  &
\end{tabular}%
\vspace{-8pt} \vspace{-4pt} \vspace{-8pt}
\caption{Cross-dataset generalization. We trained the models on the
Manhattan dataset and tested them on the London dataset. The average
geodesic error is reported.}
\label{generalization_table}
\vspace{-12pt}
\end{table}
\vspace{-8pt}
\begin{table}[tbh]
\centering
\footnotesize{
\begin{tabular}{lccccc}
\toprule \multirow{2}{*}{Overlap} & \multirow{2}{*}{Heads} & %
\multirow{2}{*}{Layers} & \multicolumn{3}{c}{Rotational error} \\
&  &  & Avg [\degree] & Med [\degree] & 10\degree [\%] \\
\toprule \multirow{5}{*}{Large} & 1 & 1 & 1.21 & 0.97 & 99.1 \\
& 4 & 1 & 0.82 & 0.65 & 99.1 \\
& 2 & 2 & 0.87 & 0.71 & 99.12 \\
& 4 & 2 & \textbf{0.58} & \textbf{0.48} & \textbf{99.31} \\
& 4 & 4 & 0.65 & 0.59 & 99.2 \\
\midrule \multirow{5}{*}{Small} & 1 & 1 & 7.13 & 3.26 & 94.99 \\
& 4 & 1 & 6.42 & 2.46 & 95.32 \\
& 2 & 2 & 5.44 & 2.05 & 96.55 \\
& 4 & 2 & \textbf{1.21} & \textbf{0.718} & \textbf{99.122} \\
& 4 & 4 & 4.61 & 0.98 & 95.43 \\
\midrule \multirow{5}{*}{None} & 1 & 1 & 7.43 & 2.88 & 91.55 \\
& 4 & 1 & 6.15 & 2.55 & 92.55 \\
& 2 & 2 & 6.23 & 3.02 & 91.35 \\
& 4 & 2 & \textbf{5.33} & \textbf{1.20} & \textbf{96.22} \\
& 4 & 4 & 5.78 & 1.87 & 95.22 \\
\bottomrule &  &  &  &  &
\end{tabular}}%
\vspace{-16pt}
\caption{Ablation of the Transformer-Encoder parameters using the StreeLearn
dataset. For each overlap class, there is an optimal configuration that
balances the Transformer-Encoder's expressive power and overfitting.}
\label{tab:ablation}
\end{table}

The cross-dataset generalization properties of our approach were evaluated
using the Holicity dataset \cite{zhou2020holicity}. The Manhattan dataset
was used to train the models, while the London dataset was used for testing.
The test images were divided into three overlap classes according to Section %
\ref{subsec:dataset}. We compared the generalization of our
approach with Cai et al.'s \cite{Cai2021Extreme}. The results, in Table ~\ref{generalization_table}, show that our
approach outperformed Cai et al in all overlap classes.

\subsection{Ablation Study}

\label{subsec:ablation}
\begin{table}[tbp]
\centering
\setlength{\tabcolsep}{1.0pt}
\footnotesize{
\begin{tabular}{cccccc}
\toprule\multirow{2}{*}{Backbone} & \multirow{2}{*}{Layers} &
\multicolumn{3}{c}{Rotational error} &  \\
& {} & Avg [\degree] & Med [\degree] & 10\degree [\%] &  \\
\midrule\multirow{4}{*}{1} & Conv(k=7, s=2, d=64) & \multirow{4}{*}{2.95} & %
\multirow{4}{*}{1.53} & \multirow{4}{*}{94.13} &  \\
& {$1 \times$ Residual blocks} &  &  &  &  \\
& {Conv(k=3, s=1, d=512)} &  &  &  &  \\
& {Conv(k=3, s=1, d=256)} &  &  &  &  \\
\midrule\multirow{4}{*}{2} & {Conv(k=7, s=2, d=64)} & \multirow{4}{*}{1.75}
& \multirow{4}{*}{1.05} & \multirow{4}{*}{96.13 } &  \\
& {$2 \times$ Residual blocks} &  &  &  &  \\
& {Conv(k=3, s=1, d=512)} &  &  &  &  \\
& {Conv(k=3, s=1, d=256)} &  &  &  &  \\
\midrule\multirow{4}{*}{3} & {Conv(k=7, s=2, d=64)} & \multirow{4}{*}{%
\textbf{1.21}} & \multirow{4}{*}{\textbf{0.7818}} & \multirow{4}{*}{%
\textbf{99.122}} &  \\
& {\textbf{3 $\times$ Residual blocks}} &  &  &  &  \\
& {Conv(k=3, s=1, d=512)} &  &  &  &  \\
& {Conv(k=3, s=1, d=256)} &  &  &  &  \\
\midrule\multirow{4}{*}{4} & {Conv(k=7, s=2, d=64)} & \multirow{4}{*}{1.45}
& \multirow{4}{*}{0.86} & \multirow{4}{*}{97.12} &  \\
& {$4 \times$ Residual blocks} &  &  &  &  \\
& {Conv(k=3, s=1, d=512)} &  &  &  &  \\
& {Conv(k=3, s=1, d=256)} &  &  &  &  \\
\bottomrule &  &  &  &  &
\end{tabular}}%
\vspace{-16pt}
\caption{Backbone Ablation. We evaluate the depth of the Residual-Unet
backbone network \protect\cite{zhang2018road}, used in our scheme, by
changing the number of residual blocks.}
\label{table:backbone-depth}
\vspace{-8pt}
\end{table}
 \textbf{Transformer-Encoder parameters.} Table \ref%
{tab:ablation} summarizes multiple Transformer-Encoder configurations for
each overlap category. The expressive power of the Transformer-Encoder
depends on the number of heads and layers. The more are used, the better the
expressive power. However, using an excessive number might lead to
overfitting, and the optimal constellation, in terms of accuracy in Table %
\ref{tab:ablation}, is given by $h=4$, $l=2$. In particular, this
constellation is a sweetspot so that increasing the number of heads or
layers results in reduced accuracy.
\begin{table}[tbp]
\centering
\setlength{\tabcolsep}{1.0pt}
\begin{tabular}{lcccc}
\toprule\multirow{2}{*}{Representation} & \multirow{2}{*}{Loss function} &
\multicolumn{3}{c}{Rotational error} \\
&  & Avg [\degree] & Med [\degree] & 10\degree [\%] \\
\midrule Quaternions & $L_{2}$ Regression & \textbf{1.21} & \textbf{0.78} &
\textbf{99.122} \\
Euler angles & Cross-Entropy & 2.37 & 1.46 & 98.13 \\
Euler angles & $L_{2}$ Regression & 2.22 & 1.32 & 98.99 \\
\bottomrule &  &  &  &
\end{tabular}%
\vspace{-8pt}
\caption{Ablation of 3D rotation encoding and training losses. We compare
the 3D rotations encodings by Euler angles and quaternions in discrete and continuous domains and the corresponding training losses.}
\label{table:regression-class}
\vspace{-16pt}
\end{table}

\textbf{Backbone ablation.} In Table \ref{table:backbone-depth}, we examine
the depth of the Residual-Unet \cite{zhang2018road} backbone by altering the
number of residual blocks it contains. Increasing the number of residual
blocks enhances the backbone's expressive capability, but an excessively
deep architecture may result in overfitting. We found that using three
residual blocks is the optimal choice, which aligns with our original
decision.

\textbf{3D rotation encoding and training losses.} The ablations of
different rotation encodings and their corresponding training losses were
evaluated and are presented in Table \ref{table:regression-class}. The
evaluation was performed by applying the Residual-Unet backbone \cite%
{zhang2018road} and a proposed Transformer-Encoder-based cross-attention
method to image pairs from the StreeLearn dataset with large overlaps. For
the discrete formulation, in line with Cai et al. \cite{Cai2021Extreme}, the
pitch and yaw angles were discretized into 360 bins $\in $ $\left[ -{%
180^{\circ },180^{\circ }}\right] $, and a cross-entropy loss was used to
train the network. These results were compared to those obtained using the $%
L_{2}$ regression loss, as described in Eq. \ref{equ:loss} in our scheme.
The results in Table \ref{table:regression-class} show that our $%
L_{2} $ regression outperforms the discrete Euler angle approach
proposed by Cai et al. \cite{Cai2021Extreme}.

\textbf{Architecture Ablations}
To evaluate the proposed architecture and assess the contribution of each
proposed component, the use of the different Transformer Decoders in
particular, we conducted a series of experiments employing various
architectural variations of the proposed architecture introduced in Section %
\ref{sub:rotation estimation} and Fig. \ref{fig:overview}. The results are
shown in Table \ref{table:arc ablation} and the corresponding architectures
are shown in the Supplementary Materials. In each experiment, we used a
particular partial configuration of the proposed Transformer Decoders and
evaluated the resulting estimation error using the StreetLearn dataset with
large overlaps between the input images. The results in Table \ref{table:arc
ablation} show that the proposed configuration outperforms all other
configurations. In particular, configuration \#1 shows that using the
sequential attention-based decoders, TD1 and TD2 improves the accuracy
significantly. The cross-decoding by TD0 provides additional, but not as
significant improvement.

\begin{table}[tbh]
\centering%
\begin{tabular}{lcccccc}
\toprule & TD0 & TD1 & TD2 & Avg [\degree] & Med [\degree] & 10\degree [\%]
\\ \hline
0 & + & + & + & \textbf{0.58} & \textbf{0.48} & \textbf{99.3} \\
1 & - & + & + & 0.98 & 0.81 & 95.15 \\
2 & - & - & - & 3.35 & 2.44 & 88.16 \\
3 & - & + & - & 1.76 & 1.55 & 93.12 \\
4 & + & + & - & 0.86 & 0.72 & 95.64 \\
5 & + & - & - & 1.97 & 1.65 & 92.82 \\
\bottomrule &  &  &  &  &  &
\end{tabular}%
\vspace{-8pt} \vspace{-8pt}
\caption{Architectural ablation study. We compare the estimation accuracy of
different configurations of Transformers Decoders (TDs). The corresponding
architectures are shown in the Supplementary Materials, and the first
configuration is shown in Section \protect\ref{sub:rotation estimation} and
Fig. \protect\ref{fig:overview}.}
\label{table:arc ablation}
\vspace{-16pt}
\end{table}

\section{Conclusion}
We present a novel formulation for estimating the relative rotation
between a pair of images. In particular, we study the estimation of
rotations between images with small and no overlap. We propose an
attention-based approach using a Transformer-Encoder to calculate the
cross-attention between image pair embedding maps, which outperforms the
previous use of 4D correlation volumes \cite{RAFT,Cai2021Extreme} and a
decoder-decoder mechanism to estimate the output quaternion. Our framework
can be trained end-to-end and optimizes a regression loss. It has been
experimentally shown to outperform previous SOTA schemes \cite%
{Cai2021Extreme} on multiple datasets used in contemporary work. In
particular, for the challenging small and nonoverlapping cases.

{\small
\bibliographystyle{ieee_fullname}
\bibliography{ref}
}

\end{document}